\documentclass[11pt]{article}
\usepackage{coling2018}
\usepackage{times}
\usepackage{url}
\usepackage{latexsym}

\usepackage{amsmath}

\title{Computing Word Classes Using Spectral Clustering}

\author{\hspace{3cm}Effi Levi  \And Saggy Herman \\
	\quad\enskip Institute of Computer Science,  The Hebrew University \\
	{\tt \{efle$|$arir\}@cs.huji.ac.il~~saggy.herman@mail.huji.ac.il} \And \hspace{-3cm}Ari Rappoport \\\\
}
\date{}

\begin{document}
\maketitle
\begin{abstract}
Clustering a lexicon of words is a well-studied problem in natural language processing (NLP). Word clusters are used to deal with sparse data in statistical language processing, as well as features for solving various NLP tasks (text categorization, question answering, named entity recognition and others).

Spectral clustering is a widely used technique in the field of image processing and speech recognition. However, it has scarcely been explored in the context of NLP; specifically, the method used in this work~\cite{meila2001random} has never been used to cluster a general word lexicon.

We apply spectral clustering to a lexicon of words, evaluating the resulting clusters by using them as features for solving two classical NLP tasks: semantic role labeling and dependency parsing. We compare performance with Brown clustering, a widely-used technique for word clustering, as well as with other clustering methods. We show that spectral clusters produce similar results to Brown clusters, and outperform other clustering methods. In addition, we quantify the overlap between spectral and Brown clusters, showing that each model captures some information which is uncaptured by the other.
\end{abstract}

\section{Introduction}
\label{sec:intro}

\emph{Word clusters} (or \emph{word classes}) are the result of dividing a lexicon of words into a pre-defined number of distinct groups, where the words comprising each group share some common characteristic. They have been well studied~\cite{martin1998algorithms}, in the context of part of speech (POS) induction~\cite{Christodoulopoulos:2010:TDU:1870658.1870714}, dealing with sparse data in statistical language processing~\cite{brown1992class}, as well as for use as features in various NLP tasks such as text categorization~\cite{bekkerman2003distributional}, question answering~\cite{momtazi2009word}, statistical parsing~\cite{candito2009improving} named entity recognition (NER)~\cite{miller2004name}, and others.

\emph{Brown Clustering}~\cite{brown1992class}, a hard hierarchical agglomerative clustering method, based on maximizing the quality of an induced class-based language model to make clustering decisions, is widely used to produce features for various NLP tasks~\cite{momtazi2009word,candito2009improving,miller2004name,koo2008simple}.

\emph{Spectral clustering} is a clustering method which is broadly used for tasks such as image segmentation~\cite{shi2000normalized,ng2002spectral}, speech recognition~\cite{bach2006learning} and topological mapping~\cite{brunskill2007topological}. 
It belongs to the family of dimensionality reduction algorithms and can actually be viewed as a weighted kernel K-means algorithm~\cite{dhillon2004kernel}.
However, it has scarcely been used for NLP tasks. ~\newcite{dhillon2011multi} use spectral methods for NER and chunking -- but not for clustering -- while \newcite{sun2009improving} employ spectral clustering to improve verb clustering. ~\newcite{sedoc2017semantic} used Signed Normalized Cut to produce word clusters; however, their motivation is focused on word-similarity (specifically, antonym-synonym relations), and they use mainly intrinsic evaluation methods. We aim to produce general spectral clusters and compare them to the widely-used Brown clusters using two NLP structured-prediction tasks for extrinsic evaluation.

While lacking the hierarchical nature of the Brown method, spectral clustering does possess a practical advantage over the former in that it produces -- in addition to word clusters -- a low-dimension representation $v \in {\rm I\!R}^m$ for each of the words in the lexicon as well as for each of the cluster centers. Beyond obvious theoretical interest, this representation may be quite useful for measuring word-word, word-cluster and even cluster-cluster distances, as well as allowing us to perform soft clustering, in which we assign each word a distribution over the clusters (instead of assigning a single cluster).

In this work we use the spectral clustering method presented in~\cite{meila2001random} over a general lexicon to produce word clusters. We proceed to show that when used as features for solving two classical yet complex NLP tasks -- SRL and dependency parsing -- spectral clusters produce very similar results to those produced by Brown clusters, despite lacking the firm language-modeling grounding inherent in the latter. Finally, we quantify the overlap (and difference) in the information contained in the two cluster sets, and show that combining them could potentially induce significantly better performance than using each on its own.

\section{Background}
\subsection{Clustering Methods}
\label{subsec:clustering}
Clustering deals with the problem of dividing set of $n$ samples into $k$ distinct clusters following a desired function measuring sample similarity. The definition of what makes two samples similar or related is not obvious or singular; most intuitively, the samples may reside in ${\rm I\!R}^n$ and the clusters should consist of samples that are similar following Euclidean distance, for example.

In the following sub-sections we will describe the best known clustering algorithm -- \emph{K-means} -- which we use as a baseline, as well as spectral clustering -- which is the focus of our work -- and Brown clustering -- which we compare to.

\subsubsection{K-means}
\label{subsubsec:kmeans}
Dating back to six decades ago and published in~\cite{lloyd1982least}, K-means is perhaps the most widely known clustering method. Given a set of points $s_1,...,s_n$ and a set of $k$ initial means $\mu_1^{(1)},...,\mu_k^{(1)}$ representing $k$ clusters $C_1^{(1)},...,C_k^{(1)}$, the algorithm iterates until convergence:

\begin{enumerate}
\item For each point, find the cluster with the closest mean and add the point to that cluster:
\begin{enumerate}
\item $i = argmin_{C^{(t)}_j}{||s_p-\mu_j^{(t)}||^2}$\\
\item $C^{(t)}_i \leftarrow s_p$
\end{enumerate}
\item Update clusters means: \\
\\
$\mu_i^{(t+1)}=\frac{1}{|C_i^{(t)}|}\sum_{s_p \in C_i^{(t)}}s_p$
\end{enumerate}

The means initialization may be random; different variants of K-means perform better with different types of initializations~\cite{hamerly2002alternatives}.

K-means performs best when the data is spatially separable, and uses Euclidean distance as metric. It also has a tendency to choose clusters occupying area of similar size on feature space.

\subsubsection{Spectral Clustering}
\label{subsubsec:spectral}
Spectral clustering refers to a group of algorithms that work on the spectral (eigenvector) decomposition of the samples' \emph{affinity matrix}, or  \emph{similarity matrix}. Each sample is represented by a vector; a metric is used in the resulting vector space to compute the distances between all point pairs -- the affinity matrix. Following some mathematical manipulation on the affinity matrix, the eigenvectors are computed, and from them the clustering is derived, usually using the K-means algorithm. There are many flavors to this method, and we will describe here two of the main algorithms. 

{\bf The first algorithm} is introduced in~\cite{ng2002spectral}:

Given a set of points $S={s_1,...,s_n}$ in ${\rm I\!R}$ that we want to cluster into $k$ subsets:
\begin{enumerate}
\item Form the affinity matrix $A \in {\rm I\!R}^{n \times n}$ defined by $A_{ij}=\exp(-||s_i-s_j||^2/2\sigma^2)$ ($A_{ii}=0$).
\item Define $D$ to be a diagonal matrix s.t. $D_{ii} = \sum_j{A_{ij}}$, and $L = D^{-1/2}AD^{-1/2}$.
\item Find the $k$ largest eigenvectors of $L$ (matching the largest $k$ eigenvalues), $x_1, \ldots ,x_k$, and stack them in columns to form $X=[x_1 \ldots x_k] \in {\rm I\!R}^{n \times k}$.
\item Derive $Y$ from $X$ by normalizing $X$'s rows to have unit length.
\item Cluster $Y$'s rows into $k$ clusters (using K-means, for example).
\item Assign point $s_i$ to cluster $j$ iff row $i$ of $Y$ was assigned to cluster $j$.
\end{enumerate}

{\bf The second algorithm} is presented by ~\newcite{shi2000normalized,meila2001random} and takes a graph-theoretic approach to clustering. The data set is represented by a weighted undirected graph. Each point is represented by a vertex, and each pair of vertices are connected by a weighted edge, with the weight representing the similarity between the two respective points. In this setting, they problem of segmenting the data set into two groups is formulated as partitioning the graph into two groups of vertices where the similarity within each group is maximized while the similarity between the groups is minimized.

Given a weighted graph $G(V,E,w)$ and two subsets of the vertices in the graph $A,B \subseteq V$, Define:
\begin{equation*}
w(A,B)=\sum_{u \in A, v \in B}w(u,v),
\end{equation*}
The normalized cut, a symmetric measure for the disassociation between the subsets, is defined to be:
\begin{equation*}
Ncut(A,B)=\frac{w(A,B)}{w(A,V)}+\frac{w(A,B)}{w(B,V)}
\end{equation*}
Along with a measure for association within each set:
\begin{equation*}
Nassoc(A,B)=\frac{w(A,A)}{w(A,V)}+\frac{w(B,B)}{w(B,V)}.
\end{equation*}
Note that these two measures are related in the following way:
\begin{equation*}
Ncut(A,B)=2-Nassoc(A,B)
\end{equation*}
Therefore, by minimizing the disassociation between groups we also maximize the association within each group.

Finding a cut that minimizes the \emph{Ncut} criterion is shown to be NP-hard, and the \emph{Ncut algorithm} is introduced, approximating in polynomial time the 2-way cut solution using the eigenvalues and eigenvectors of the affinity matrix. The algorithm is then used recursively to find a $k$-way partition of the graph, providing a clustering of the data set to $k$ groups.

A more efficient method, however, is presented in~\cite{meila2001random}. After establishing a probabilistic theoretic foundation to the normalized cut framework by offering a random walk interpretation, they present the \emph{Modified Ncut algorithm} for a one-pass $k$-way segmentation:
\begin{enumerate}
\item Generate $D$ as defined in~\cite{ng2002spectral} (the first algorithm).
\item Generate $P=D^{-1}S$, $S$ being the similarity matrix, and find its eigenvalues and eigenvectors.
\item Discard the leading eigenvector and stack second-through-$k$ leading eigenvectors to form $X=[x_2 \ldots x_k]$.
\item Perform K-means (or equivalent) on rows of $X$ to find the clusters.
\end{enumerate}

We chose to use the latter algorithm in our work, due to availability-of-code considerations.

Note that the last stage in the algorithm involves clustering the data points in the dense (due to the dimensionality reduction) vector space created by the computed eigenvectors. As mentioned in Section~\ref{sec:intro}, this means that a by-product of this algorithm is a vector representation in this space for each of the clustered data points (in our case, lexicon words) as well as for the cluster centers. We discuss this further in Section~\ref{sec:conclusion}.

\subsubsection{Brown Clustering}
\label{subsub:brown}
Brown clustering~\cite{brown1992class} is a hard hierarchical agglomerative clustering method. It is based on the concept of maximizing the quality of an induced class-based language model, having originally been presented as a class-based n-gram model for natural language. 

The algorithm generally follows the outline for hard hierarchical agglomerative clustering:
\begin{enumerate}
\item Start with a lexicon of types $V$
\item Sort $V$ by corpus frequency, then put the top $k$ types into clusters $C_1,...,C_k$
\item Repeat $|V| - k$ times: 
\begin{itemize}
\item Put the next type into a new cluster $C_{k+1}$
\item Merge the pair in the $k+1$ clusters to receive clustering $C$ that maximizes $Quality(C)$
\end{itemize}
\end{enumerate}

The function measuring the quality of the clustering at each iteration -- $Quality(C)$ -- was defined based on statistical language modeling reasoning. 
Given a corpus $w_1,...,w_n$, a first-order Hidden Markov Model is used (with the clusters $C(w_1),...,C(w_n)$ as the latent variables) to approximate the corpus probability:
\begin{equation*}
\begin{aligned}
P&(w_1,..,w_n|C(w_1),...,C(w_n)) \approx \\ 
& \prod_{i=1}^{n}{P(w_i | C(w_i))P(C(w_i)|C(w_{i-1}))}
\end{aligned}
\end{equation*}

The quality function is defined to be:
\begin{equation*}
\small
\begin{aligned}
Qua&lity(C) = \frac{1}{n}\log{P(w_1,..,w_n|C(w_1),...,C(w_n))} \\ 
& \approx \frac{1}{n}\log{\prod_{i=1}^{n}{P(w_i | C(w_i))P(C(w_i)|C(w_{i-1}))}}
\end{aligned}
\end{equation*}

Further decomposing into:
\begin{equation*}
\begin{aligned}
&= \sum_{1 \leq i,j \leq k}{P(C_i,C_j)\log{\frac{P(C_i,C_j)}{P(C_i)P(C_j)}}} \\
& \qquad \qquad + \sum_w{P(w)\log{P(w)}} \\
&= I(C) - H(V)
\end{aligned}
\end{equation*}

Which is the mutual information of the clustering minus the entropy of the vocabulary. The latter is constant over the clustering, meaning that the mutual information is maximized over the clustering in each iteration of the algorithm.

\subsection{Evaluation}
\label{sub:eval}
Conducting a comparison between two different clustering methods is not a trivial task. Viewing the clusters themselves may provide us with some insight. Some criteria exist for assessing the quality of the clusters, such as silhouette graphs~\cite{rousseeuw1987silhouettes}, while others, such as the Variation of Information criterion(VI) ~\cite{meilua2007comparing} along with its normalized variant~\cite{Reichart:2009:NCE:1596374.1596401}, intend to compare two different clusterings over the same data set. These intrinsic methods, however, offer us very limited insight due to the unsupervised nature of our clustering task. We therefore turn to an extrinsic method of evaluation: using the clusters as features in a higher level task, and evaluate the clusters based on the performance in that task. We accomplish this by testing the clusters as features in two classical structured-prediction NLP tasks: semantic role labeling (SRL) and dependency parsing. As we wish to test the power of word clusters as features for these tasks, we use them exclusively without using any syntactic (POS tagging or dependency parses) or other semantic information.

\subsubsection{Semantic Role Labeling}
\label{subsub:srl}
Semantic role labeling (SRL) is the task of detecting and labeling the different semantic arguments of a predicate in a sentence. The foundations for the task have been laid in~\cite{gildea2002automatic}, and had attracted much attention since. We use the PropBank annotated WSJ corpus, ~\newcite{palmer2005proposition}, which expands the Penn TreeBank with semantic annotations. These annotations were used as gold standard in the CoNLL-2005 shared task~\cite{carreras2005introduction}.

While most often the task is broken down to two parts --- finding the arguments using syntactic features, then labelling them using semantic features~\cite{marquez2008semantic} -- others try to solve the problem holistically, usually by training a statistical model such as Hidden Markov Model (HMM) or Conditional Random Fields (CRF), often solving some syntactic task jointly with SRL~\cite{henderson2008latent}. We adopt this approach in our work and use CRF for performing SRL.

Very few works attempt SRL without any syntactic features; ~\newcite{boxwell2011semantic} report $F_1= 0.44$ (although it is only reported for ``completeness''). Recently, neural networks models have shown great promise in solving SRL. ~\newcite{boxwell2011semantic}, employing a unified neural network model to jointly learn POS tagging, chunking, NER and SRL, report $F_1 = 0.74$ on the SRL task. ~\newcite{zhou2015end}, using a bi-directional long short-term memory (LSTM), report $F_1 = 81.07$, making their system state-of-the-art. These systems, however, are very complex and embody extensive fine-tuning in order to achieve the best possible results on SRL; our motivation, as previously stated, lies elsewhere.

\subsubsection{Dependency Parsing}
\label{subsub:dep}
Dependency parsers build on the syntactic theory of \emph{Dependency Grammar}. Proposed by Lucien Tesni\`{e}re (1893-1954), the theory is based on relationships between words --- between a \emph{head} and a \emph{dependent}. Starting from the verb, directed links connect all words of the sentence with links pointing from head to dependent, creating a directed rooted tree.

Dependency parsing is a major component of a large number of NLP applications. It is therefore one of the most well-studied tasks in NLP~\cite{mcdonald2005non,nivre2007maltparser,zhang2011transition} and has been the focus of the CoNLL-2007 shared task~\cite{nilsson2007conll}. Extending classic syntactic features, ~\newcite{koo2008simple} use 4-6 bit prefixes of Brown clusters and full length clusters along with POS tags and word forms. They report an improvement in accuracy to 93.16\%. over a baseline of 92.02\%. ~\newcite{bansal2014tailoring} challenge Brown clusters by using word embeddings and performing hierarchical clustering on them. They reach accuracy of 92.69\%, same as their Brown baseline, but with considerably faster training time.

\section{Experimental Setup}
\label{sec:expset}

\subsection{Representation}
\label{subsec:rep}
in order to employ K-means over the lexicon as well as compute an affinity matrix for spectral clustering, we first need to decide on a representation for the words in our lexicon. We choose to use a simple window-based count model, in which each word is represented by the number of times it appears within a window (with a pre-defined size) around a pre-defined set of \emph{descriptor words} in some corpus. 

More formally, given a corpus, we choose the set of $m$ most frequent words in the corpus to be our descriptor words, then each word $w$ in the corpus is represented as a vector in ${\rm I\!R}^M$, where each coordinate denotes the number of times $w$ appears in the corpus in a window of a pre-determined size $W$ on the {\bf right} size of the respective descriptor word. The result is a $N \times M$ matrix (where $N$ is the number of word in our lexicon) denoted $R$. We compute the exact same matrix using the {\bf left} size of the descriptor word, and denote it $L$. These two matrices are finally concatenated to create a $N \times 2M$ sized matrix, in which each row contains the vector representation of a lexicon word, denoted the \emph{context matrix} $C$.

In order to generate the context matrix we use the ukWaC corpus~\cite{baroni2009wacky}, containing 2 billion words crawled from the~\url{.uk} domain. Words are tokenized following the CoNLL format (separated \texttt{'s}, \texttt{'nt}, etc.), and the first word of a sentence is decapitalized.

For practical reasons, we artificially limit our lexicon size by choosing the top $N-1$ most frequent words in the corpus, treating all other words as a special token "RARE" (making it the $N$-th word in our lexicon).

For the experiments in this work we use $N=12007$ and $M=5000$ (the values were chosen empirically). In addition, we experiment with various values for the window size $W$ (2, 3 and 5).

In addition, for completeness, we perform an additional set of experiments using a state-of-the-art word embedding as our lexicon words representation. We chose to use \texttt{word2vec}, implementing the continuous skip-gram algorithm presented in ~\newcite{mikolov2013efficient} with negative sampling~\cite{mikolov2013distributed}.

\subsection{Clustering}
\subsubsection{Number of Clusters} 
We set the number of clusters on $k=250$ throughout our experiments for all the clustering methods used in this work. This value was chosen empirically, having proven to provide the best results for both spectral and Brown clusters on the SRL task in preliminary experiments.

\subsubsection{Affinity Matrix}
As noted in Section~\ref{subsubsec:spectral}, the spectral clustering algorithm required an affinity matrix, representing the affinity between each pair of samples (lexicon words) as input. Rather than using the default Gaussian kernel~\cite{ng2002spectral}, we follow~\newcite{sun2009improving} using the \emph{symmetrized skew-divergence} for generating the affinity matrix. This approach was found to produce better results during preliminary testing. Given two vectors $v$ and $v'$, their skew-divergence is
\begin{equation*}
d_{skew}(v,v')=D_{KL}(v'||a \cdot v+(1-a) \cdot v')
\end{equation*}
Where $D_{KL}$ is the KL-divergence and $v$ is smoothed with $v'$ by parameter $a$ (we empirically choose $a=0.999$). The symmetrized skew-divergence is then defined as:
\begin{equation*}
d_{s-skew}(v,v')=\dfrac{1}{2}(d_{skew}(v,v')+d_{skew}(v',v))
\end{equation*}

Finally, the affinity matrix is computed using this measure. Given the $i$-th and $j$-th lexicon words ($i \neq j$), we compute their respective vector representations $v_i,v_j$. We then set
\begin{equation*}
A_{ij} = A_{ji} = d_{s-skew}(v,v')
\end{equation*}

For each $i$, we set $A_{ii} = 0$.

\subsection{Software}
\subsubsection{Word Embeddings}  
In some of our experiments we utilize the \texttt{word2vec} embedding as word representation. We use the word embeddings available at \url{https://code.google.com/p/word2vec/}. These embeddings were produced by a network which was trained on a partial Google News data set ($\sim$100 billion words), and generated 300-dimensional vectors.

\subsubsection{K-means} 
The K-means algorithm is employed as a baseline, as well as for the final stage in the spectral clustering algorithm (see Section~\ref{subsubsec:spectral}). We use \textsc{matlab}'s implementation of Lloyd's algorithm~\cite{lloyd1982least}, with the \textsc{k-means++} algorithm for centroid initialization~\cite{arthur2007k}.

\subsubsection{Spectral Clustering} 
We use the spectral clustering package presented in~\cite{cour2004normalized}, implementing the NCut spectral segmentation algorithm presented in \cite{shi2000normalized}, available at~\url{http://www.timotheecour.com/software/ncut/ncut.html}.

\subsubsection{Brown Clustering} 
We use the implementation provided by~\newcite{liang2005semi}, available at~\url{https://github.com/percyliang/brown-cluster}. The clusters were computed on the ukWaC corpus described in Section~\ref{subsec:rep}.

\subsubsection{CRF} 
For the purpose of solving SRL we wanted a simple yet powerful, off-the-shelf learning algorithm. As stated in Section~\ref{subsub:srl}, we choose to use CRF, feeling it is a simple yet powerful enough tool that would allow us to place the focus on the features rather than the learning process. We use the \textsc{CRF++} package, available at~\url{https://taku910.github.io/crfpp/}, implementing~\newcite{lafferty2001conditional}'s algorithm.

\subsubsection{Dependency Parser}
For our dependency parsing experiments we utilized the MSTParser, implementing the parser described in~\cite{mcdonald2005non} supplemented with second order features~\cite{mcdonald2006online}, available at~\url{http://sourceforge.net/projects/mstparser/}.

\section{Experiments \& Results}
\label{sec:exp}
\subsection{SRL}
\label{subsec:exp_srl}
Our first set of experiments is conducted on the PropBank annotated WSJ corpus~\cite{palmer2005proposition}, which expands the Penn TreeBank with semantic annotations. We train a CRF model on sections 2-21 and test it on section 23. The features set for the CRF algorithm includes five surface features: whether the word contains a number, a hyphen or a capital letter, the position of the word w.r.t. current predicate and the word's length in characters. In each experiment we add one of the following sets of features:

\begin{enumerate}
\item K-means clusters (using simple count model decribed in~\ref{subsec:rep})
\item Spectral clusters (using simple count model decribed in~\ref{subsec:rep})~\footnote{Using a combination of all three window sizes, discussed in Section~\ref{subsec:rep}}
\item Brown clusters
\item K-means clusters (using \texttt{word2vec})
\item Spectral clusters (using \texttt{word2vec})
\item POS tag
\item POS tag + dependency~\footnote{\label{dep_note}Dependency parses were extracted from the CoNLL-2005 Shared Task annotated data set~\cite{carreras2005introduction}.} parent's POS tag
\item POS tag + dependency~\footnotemark[2] parent's POS tag + dependency~\footnotemark[2] grandparent's POS tag
\end{enumerate}

For each experiment we report per-argument precision, recall and $F_1$ score (Table~\ref{tab:srl_results}).

\begin{table*}[t!]
\centering
\begin{tabular}{lclclclcl}
\hline
{\bf Feature Set}     					    & {\bf precision} & {\bf recall}  & \boldmath$F_1$   \\
\hline
K-means (count model)                                  & 0.620 & 0.504 & 0.555 \\
Spectral (count model)                                   & 0.648 & 0.552 & 0.596 \\
Brown                                                    	    & 0.663 & 0.547 & 0.599 \\
K-means (\texttt{word2vec})                           & 0.628 & 0.471 & 0.539 \\
Spectral (\texttt{word2vec})                            & 0.642 & 0.503 & 0.565 \\
POS                                              		    & 0.666 & 0.552 & 0.604 \\	
POS + parent's POS                               	    & 0.661 & 0.556 & 0.604 \\
POS + parent's POS + grandparent's POS    & 0.681 & 0.564 & 0.617 \\
\hline
\end{tabular}
\caption{Results for the SRL experiments.}
\label{tab:srl_results}
\end{table*}

Examining the first 3 rows in Table~\ref{tab:srl_results}, we note the very small difference in $F_1$ score between the spectral model and the Brown model (0.003), denoting virtually equal performance. This is a very interesting result, considering that Brown clustering is tailored for word clustering by incorporating a statistical language model in the target function used for the clustering itself, as opposed to spectral clustering which takes no lexical considerations into account during the clustering process. Both models improve over the baseline (K-means) by approximately 0.04.

Moving down in the table, we see that experimenting with the \texttt{word2vec} embedding as initial representation yields worse results than using the simple count model. This is also a surprising result given the amount of success \texttt{word2vec} is having in word similarity tasks~\cite{baroni2014don}.
We suspect this is due to the fact that these embeddings are learned without any connection to the clustering task, possibly not retaining some clustering-related information due to their high density.

Moving on, we observe that using POS tags as features instead of word clusters does not significantly improve the results: merely by 0.005 over Brown clusters. Considering the POS tag of the word's parent in the dependency tree does not improve the results, but considering the POS tag of the grandparent does improve performance by 0.013. All-in-all, we can improve over word clusters using syntactic information by the small amount of 0.018, but this requires second-order information from the dependency tree as well as POS tags, both which are generally expensive to manually produce.

\subsection{Dependency Parsing}
\label{subsec:exp_dep}
Our second set of experiments is conducted on the dependency annotations expansion to the WSJ corpus introduced in~\cite{carreras2005introduction}. We train the MSTParser on sections 2-21 of the data set and test the resulting model on section 23. We use several of the feature sets described in Section~\ref{subsec:exp_srl} to compare between the different clustering techniques. 

For each experiment we report unlabeled attachment score (UAS). Results are shown in Table~\ref{tab:dep_results}.

\begin{table*}[t!]
\centering
\begin{tabular}{lclcl}
\hline
{\bf Feature Set}     					    & {\bf UAS} \\
\hline
Spectral (count model)                                   & 0.867 \\
Brown                                                    	    & 0.881 \\
POS                                              		    & 0.921\\	
\hline
\end{tabular}
\caption{Results for the dependency parsing experiments.}
\label{tab:dep_results}
\end{table*}

We can observe that these results are similar to the ones obtained for SRL. The difference between using Brown clusters to spectral ones is relatively small, though larger than in the SRL task (0.014 out of 0.881). Using POS improves performance here too, though by a larger margin (0.041). Over all, we observe the same phenomena, albeit on a slightly different scale.

\subsection{Oracle Model}
In order to analyze the amount of information overlap between spectral and Brown clusters (in the context of SRL and dependency parsing performance), we examine the performance of an \emph{oracle model}.

Given a prediction task, along with two models trained for the task, an oracle model is a hypothetical model which is capable of determining which one of the two models will perform better on any given test sample. The decision is made per sample; for each sample, the oracle chooses the better model. As stated, this is obviously a hypothetical model, but it gives us the opportunity to estimate how overlapping our two models are. In the case of a complete overlap, the performance of the oracle model will not improve beyond those of the separate models; if the overlap is not complete, however, the amount of improvement achieved by the oracle may indicate the amount of difference between the models.

We perform the following procedure for the models learned using the spectral vs. the Brown clusters (feature sets 2 \& 3 in~\ref{subsec:exp_srl}). We go over the test samples one by one and check which model performed better, taking that model's prediction to be the oracle's prediction on that test sample. Finally, we evaluate the oracle model according to its predictions. We perform this analysis for both the SRL and the dependency parsing data sets. 

Results -- $F_1$ score for SRL and UAS for dependency parsing -- are given in Table~\ref{tab:oracle_results}.

\begin{table*}[t!]
\centering
\begin{tabular}{lclclc|c|}
\hline
{\bf Task}	&  {\bf Spectral}   & {\bf Brown}    & {\bf Oracle} \\
\hline
SRL ($F_1$)   &	0.596	&	0.599	&	0.664 \\
\hline
Dep. Parsing (UAS)	&	0.867	&	0.881	&	0.901 \\
\hline
\end{tabular}
\caption{Results for oracle model analysis.}
\label{tab:oracle_results}
\end{table*}

The oracle result for SRL reveals the complementary nature of the two methods, achieving a significant increase of 0.075-0.078 in $F_1$ score over using each model separately (an improvement of 12.5\%) , outperforming even the best model which is based on syntactic features by 0.047 in $F_1$ score (compare to last row in Table~\ref{tab:dep_results}). This interesting result means that the information captured by the spectral method may be quite different from the one captured by the Brown clusters.

A deeper examination reveals that while the two models agree on 61.4\% of the sentences in the test set, the spectral model outperforms the Brown model on 18.8\% of the sentences while the Brown model performs better on 19.8\% of them, further emphasizing the complementary nature of the two clustering methods.

The oracle result for dependency parsing shows an improvement over the individual models as well, although a less dramatic one (an increase of 0.02-0.034 in UAS, which is a 2.3\%-3.9\% improvement). We hypothesize that the reason may be the syntactic nature of dependency parsing compared to the semantic nature of SRL, given that distributional word clusterings are usually assumed to capture mainly semantic information.

\section{Conclusion}
\label{sec:conclusion}
In this work we apply spectral clustering to produce word clusters for a general word lexicon. We compare them to a K-means baseline, as well as the acclaimed Brown clusters, by using the clusters resulting form each of these methods as feature for solving SRL and dependency parsing. For a complete comparison, we also examine the use of an advanced word embedding method (\texttt{word2vec}) as well as hand-crafted syntax-based features (POS tags, dependency trees) in the same setup.

Interestingly, we observe a very similar (virtually identical, in the case of SRL) performance by the spectral and the Brown clusters, both outperforming the other clustering methods, while being marginally outperformed by hand-crafted syntactic features.  We find the similar performance exhibited by the spectral clusters, compared to Brown clusters, a very interesting result, given the linguistically-motivated nature of Brown clusters vs.\ the none-language-related spectral clustering method.

In our view, this result may serve to motivate the use of the spectral method for lexicon clustering. This motivation is further enhanced by an advantageous quality of this clustering method: as a by-product, it produces a low-dimension vector representation for each of the lexicon words as well as for each of the clusters themselves (see the end of Section~\ref{subsubsec:spectral}). This representation may be used, for example, to produce more features, such as the $i$-th closest cluster for each word, or produce a \emph{soft} clustering, assigning each word with the distribution over distances from the clusters. It may also be used it to remove word outliers from the clusters, or characterize the relations between the clusters (based on the distances between them). It may even be interesting to explore the properties of this representation as a word embedding for various NLP tasks.

An oracle analysis reveals that spectral clusters and Brown clusters complement each other rather than completely overlap. The oracle model achieves a significant improvement of 12.5\% in performance in the SRL task, with a more modest improvement of 2.3\%-3.9\% in the dependency parsing task. We further show that the two models agree merely on 61.4\% of the test samples in the SRL test set, approximately evenly dividing the rest of samples between them (in terms of besting each other). This analysis reveals the complementary nature of these clustering methods, implying that each of them may be better suited for different cases in the same task. 

\bibliographystyle{acl}
\bibliography{arxiv}

\begin{thebibliography}{}

\bibitem[\protect\citename{Arthur and Vassilvitskii}2007]{arthur2007k}
David Arthur and Sergei Vassilvitskii.
\newblock 2007.
\newblock k-means++: The advantages of careful seeding.
\newblock In {\em Proceedings of the eighteenth annual ACM-SIAM symposium on
  Discrete algorithms}, pages 1027--1035. Society for Industrial and Applied
  Mathematics.

\bibitem[\protect\citename{Bach and Jordan}2006]{bach2006learning}
Francis~R Bach and Michael~I Jordan.
\newblock 2006.
\newblock Learning spectral clustering, with application to speech separation.
\newblock {\em Journal of Machine Learning Research}, 7(Oct):1963--2001.

\bibitem[\protect\citename{Bansal \bgroup et al.\egroup
  }2014]{bansal2014tailoring}
Mohit Bansal, Kevin Gimpel, and Karen Livescu.
\newblock 2014.
\newblock Tailoring continuous word representations for dependency parsing.
\newblock In {\em ACL (2)}, pages 809--815.

\bibitem[\protect\citename{Baroni \bgroup et al.\egroup }2009]{baroni2009wacky}
Marco Baroni, Silvia Bernardini, Adriano Ferraresi, and Eros Zanchetta.
\newblock 2009.
\newblock The wacky wide web: a collection of very large linguistically
  processed web-crawled corpora.
\newblock {\em Language resources and evaluation}, 43(3):209--226.

\bibitem[\protect\citename{Baroni \bgroup et al.\egroup }2014]{baroni2014don}
Marco Baroni, Georgiana Dinu, and Germ{\'a}n Kruszewski.
\newblock 2014.
\newblock Dont count, predict! a systematic comparison of context-counting vs.
  context-predicting semantic vectors.
\newblock In {\em Proceedings of the 52nd Annual Meeting of the Association for
  Computational Linguistics}, volume~1, pages 238--247.

\bibitem[\protect\citename{Bekkerman \bgroup et al.\egroup
  }2003]{bekkerman2003distributional}
Ron Bekkerman, Ran El-Yaniv, Naftali Tishby, and Yoad Winter.
\newblock 2003.
\newblock Distributional word clusters vs. words for text categorization.
\newblock {\em Journal of Machine Learning Research}, 3(Mar):1183--1208.

\bibitem[\protect\citename{Boxwell \bgroup et al.\egroup
  }2011]{boxwell2011semantic}
Stephen~A Boxwell, Chris Brew, Jason Baldridge, Dennis Mehay, and Sujith Ravi.
\newblock 2011.
\newblock Semantic role labeling without treebanks?
\newblock In {\em IJCNLP}, pages 192--200.

\bibitem[\protect\citename{Brown \bgroup et al.\egroup }1992]{brown1992class}
Peter~F Brown, Peter~V Desouza, Robert~L Mercer, Vincent J~Della Pietra, and
  Jenifer~C Lai.
\newblock 1992.
\newblock Class-based n-gram models of natural language.
\newblock {\em Computational linguistics}, 18(4):467--479.

\bibitem[\protect\citename{Brunskill \bgroup et al.\egroup
  }2007]{brunskill2007topological}
Emma Brunskill, Thomas Kollar, and Nicholas Roy.
\newblock 2007.
\newblock Topological mapping using spectral clustering and classification.
\newblock In {\em Intelligent Robots and Systems, 2007. IROS 2007. IEEE/RSJ
  International Conference on}, pages 3491--3496. IEEE.

\bibitem[\protect\citename{Candito and Crabb{\'e}}2009]{candito2009improving}
Marie Candito and Beno{\^\i}t Crabb{\'e}.
\newblock 2009.
\newblock Improving generative statistical parsing with semi-supervised word
  clustering.
\newblock In {\em Proceedings of the 11th International Conference on Parsing
  Technologies}, pages 138--141. Association for Computational Linguistics.

\bibitem[\protect\citename{Carreras and
  M{\`a}rquez}2005]{carreras2005introduction}
Xavier Carreras and Llu{\'\i}s M{\`a}rquez.
\newblock 2005.
\newblock Introduction to the conll-2005 shared task: Semantic role labeling.
\newblock In {\em Proceedings of the Ninth Conference on Computational Natural
  Language Learning}, pages 152--164. Association for Computational
  Linguistics.

\bibitem[\protect\citename{Christodoulopoulos \bgroup et al.\egroup
  }2010]{Christodoulopoulos:2010:TDU:1870658.1870714}
Christos Christodoulopoulos, Sharon Goldwater, and Mark Steedman.
\newblock 2010.
\newblock Two decades of unsupervised pos induction: How far have we come?
\newblock In {\em Proceedings of the 2010 Conference on Empirical Methods in
  Natural Language Processing}, EMNLP '10, pages 575--584, Stroudsburg, PA,
  USA. Association for Computational Linguistics.

\bibitem[\protect\citename{Cour \bgroup et al.\egroup
  }2004]{cour2004normalized}
Timothee Cour, Stella Yu, and Jianbo Shi.
\newblock 2004.
\newblock Normalized cut segmentation code. copyright 2004 university of
  pennsylvania.
\newblock {\em Computer and Information Science Department}.

\bibitem[\protect\citename{Dhillon \bgroup et al.\egroup
  }2004]{dhillon2004kernel}
Inderjit~S Dhillon, Yuqiang Guan, and Brian Kulis.
\newblock 2004.
\newblock Kernel k-means: spectral clustering and normalized cuts.
\newblock In {\em Proceedings of the tenth ACM SIGKDD international conference
  on Knowledge discovery and data mining}, pages 551--556. ACM.

\bibitem[\protect\citename{Dhillon \bgroup et al.\egroup
  }2011]{dhillon2011multi}
Paramveer Dhillon, Dean~P Foster, and Lyle~H Ungar.
\newblock 2011.
\newblock Multi-view learning of word embeddings via cca.
\newblock In {\em Advances in Neural Information Processing Systems}, pages
  199--207.

\bibitem[\protect\citename{Gildea and Jurafsky}2002]{gildea2002automatic}
Daniel Gildea and Daniel Jurafsky.
\newblock 2002.
\newblock Automatic labeling of semantic roles.
\newblock {\em Computational linguistics}, 28(3):245--288.

\bibitem[\protect\citename{Hamerly and Elkan}2002]{hamerly2002alternatives}
Greg Hamerly and Charles Elkan.
\newblock 2002.
\newblock Alternatives to the k-means algorithm that find better clusterings.
\newblock In {\em Proceedings of the eleventh international conference on
  Information and knowledge management}, pages 600--607. ACM.

\bibitem[\protect\citename{Henderson \bgroup et al.\egroup
  }2008]{henderson2008latent}
James Henderson, Paola Merlo, Gabriele Musillo, and Ivan Titov.
\newblock 2008.
\newblock A latent variable model of synchronous parsing for syntactic and
  semantic dependencies.
\newblock In {\em Proceedings of the Twelfth Conference on Computational
  Natural Language Learning}, pages 178--182. Association for Computational
  Linguistics.

\bibitem[\protect\citename{Koo \bgroup et al.\egroup }2008]{koo2008simple}
Terry Koo, Xavier Carreras~P{\'e}rez, and Michael Collins.
\newblock 2008.
\newblock Simple semi-supervised dependency parsing.
\newblock In {\em 46th Annual Meeting of the Association for Computational
  Linguistics}, pages 595--603.

\bibitem[\protect\citename{Lafferty \bgroup et al.\egroup
  }2001]{lafferty2001conditional}
John~D. Lafferty, Andrew McCallum, and Fernando C.~N. Pereira.
\newblock 2001.
\newblock Conditional random fields: Probabilistic models for segmenting and
  labeling sequence data.
\newblock In {\em Proceedings of the Eighteenth International Conference on
  Machine Learning}, ICML '01, pages 282--289, San Francisco, CA, USA. Morgan
  Kaufmann Publishers Inc.

\bibitem[\protect\citename{Liang}2005]{liang2005semi}
Percy Liang.
\newblock 2005.
\newblock {\em Semi-supervised learning for natural language}.
\newblock {Ph.D.} thesis, Massachusetts Institute of Technology.

\bibitem[\protect\citename{Lloyd}1982]{lloyd1982least}
Stuart~P Lloyd.
\newblock 1982.
\newblock Least squares quantization in pcm.
\newblock {\em Information Theory, IEEE Transactions on}, 28(2):129--137.

\bibitem[\protect\citename{M{\`a}rquez \bgroup et al.\egroup
  }2008]{marquez2008semantic}
Llu{\'\i}s M{\`a}rquez, Xavier Carreras, Kenneth~C Litkowski, and Suzanne
  Stevenson.
\newblock 2008.
\newblock Semantic role labeling: an introduction to the special issue.
\newblock {\em Computational linguistics}, 34(2):145--159.

\bibitem[\protect\citename{Martin \bgroup et al.\egroup
  }1998]{martin1998algorithms}
Sven Martin, J{\"o}rg Liermann, and Hermann Ney.
\newblock 1998.
\newblock Algorithms for bigram and trigram word clustering.
\newblock {\em Speech communication}, 24(1):19--37.

\bibitem[\protect\citename{McDonald and Pereira}2006]{mcdonald2006online}
Ryan~T McDonald and Fernando~CN Pereira.
\newblock 2006.
\newblock Online learning of approximate dependency parsing algorithms.
\newblock In {\em EACL}.

\bibitem[\protect\citename{McDonald \bgroup et al.\egroup
  }2005]{mcdonald2005non}
Ryan McDonald, Fernando Pereira, Kiril Ribarov, and Jan Haji{\v{c}}.
\newblock 2005.
\newblock Non-projective dependency parsing using spanning tree algorithms.
\newblock In {\em Proceedings of the conference on Human Language Technology
  and Empirical Methods in Natural Language Processing}, pages 523--530.
  Association for Computational Linguistics.

\bibitem[\protect\citename{Meila and Shi}2001]{meila2001random}
Marina Meila and Jianbo Shi.
\newblock 2001.
\newblock A random walks view of spectral segmentation.
\newblock {\em AI and STATISTICS (AISTATS) 2001}.

\bibitem[\protect\citename{Meil{\u{a}}}2007]{meilua2007comparing}
Marina Meil{\u{a}}.
\newblock 2007.
\newblock Comparing clusterings\textemdash an information based distance.
\newblock {\em Journal of multivariate analysis}, 98(5):873--895.

\bibitem[\protect\citename{Mikolov \bgroup et al.\egroup
  }2013a]{mikolov2013efficient}
Tomas Mikolov, Kai Chen, Greg Corrado, and Jeffrey Dean.
\newblock 2013a.
\newblock Efficient estimation of word representations in vector space.
\newblock {\em arXiv preprint arXiv:1301.3781}.

\bibitem[\protect\citename{Mikolov \bgroup et al.\egroup
  }2013b]{mikolov2013distributed}
Tomas Mikolov, Ilya Sutskever, Kai Chen, Greg~S Corrado, and Jeff Dean.
\newblock 2013b.
\newblock Distributed representations of words and phrases and their
  compositionality.
\newblock In {\em Advances in neural information processing systems}, pages
  3111--3119.

\bibitem[\protect\citename{Miller \bgroup et al.\egroup }2004]{miller2004name}
Scott Miller, Jethran Guinness, and Alex Zamanian.
\newblock 2004.
\newblock Name tagging with word clusters and discriminative training.
\newblock In {\em HLT-NAACL}, volume~4, pages 337--342.

\bibitem[\protect\citename{Momtazi and Klakow}2009]{momtazi2009word}
Saeedeh Momtazi and Dietrich Klakow.
\newblock 2009.
\newblock A word clustering approach for language model-based sentence
  retrieval in question answering systems.
\newblock In {\em Proceedings of the 18th ACM conference on Information and
  knowledge management}, pages 1911--1914. ACM.

\bibitem[\protect\citename{Ng \bgroup et al.\egroup }2002]{ng2002spectral}
Andrew~Y Ng, Michael~I Jordan, Yair Weiss, et~al.
\newblock 2002.
\newblock On spectral clustering: Analysis and an algorithm.
\newblock {\em Advances in neural information processing systems}, 2:849--856.

\bibitem[\protect\citename{Nilsson \bgroup et al.\egroup
  }2007]{nilsson2007conll}
Jens Nilsson, Sebastian Riedel, and Deniz Yuret.
\newblock 2007.
\newblock The conll 2007 shared task on dependency parsing.
\newblock In {\em Proceedings of the CoNLL shared task session of EMNLP-CoNLL},
  pages 915--932. sn.

\bibitem[\protect\citename{Nivre \bgroup et al.\egroup
  }2007]{nivre2007maltparser}
Joakim Nivre, Johan Hall, Jens Nilsson, Atanas Chanev, G{\"u}lsen Eryigit,
  Sandra K{\"u}bler, Svetoslav Marinov, and Erwin Marsi.
\newblock 2007.
\newblock Maltparser: A language-independent system for data-driven dependency
  parsing.
\newblock {\em Natural Language Engineering}, 13(02):95--135.

\bibitem[\protect\citename{Palmer \bgroup et al.\egroup
  }2005]{palmer2005proposition}
Martha Palmer, Daniel Gildea, and Paul Kingsbury.
\newblock 2005.
\newblock The proposition bank: An annotated corpus of semantic roles.
\newblock {\em Computational linguistics}, 31(1):71--106.

\bibitem[\protect\citename{Reichart and
  Rappoport}2009]{Reichart:2009:NCE:1596374.1596401}
Roi Reichart and Ari Rappoport.
\newblock 2009.
\newblock The nvi clustering evaluation measure.
\newblock In {\em Proceedings of the Thirteenth Conference on Computational
  Natural Language Learning}, CoNLL '09, pages 165--173, Stroudsburg, PA, USA.
  Association for Computational Linguistics.

\bibitem[\protect\citename{Rousseeuw}1987]{rousseeuw1987silhouettes}
Peter~J Rousseeuw.
\newblock 1987.
\newblock Silhouettes: a graphical aid to the interpretation and validation of
  cluster analysis.
\newblock {\em Journal of computational and applied mathematics}, 20:53--65.

\bibitem[\protect\citename{Sedoc \bgroup et al.\egroup
  }2017]{sedoc2017semantic}
Joao Sedoc, Jean Gallier, Dean Foster, and Lyle Ungar.
\newblock 2017.
\newblock Semantic word clusters using signed spectral clustering.
\newblock In {\em Proceedings of the 55th Annual Meeting of the Association for
  Computational Linguistics (Volume 1: Long Papers)}, volume~1, pages 939--949.

\bibitem[\protect\citename{Shi and Malik}2000]{shi2000normalized}
Jianbo Shi and Jitendra Malik.
\newblock 2000.
\newblock Normalized cuts and image segmentation.
\newblock {\em Pattern Analysis and Machine Intelligence, IEEE Transactions
  on}, 22(8):888--905.

\bibitem[\protect\citename{Sun and Korhonen}2009]{sun2009improving}
Lin Sun and Anna Korhonen.
\newblock 2009.
\newblock Improving verb clustering with automatically acquired selectional
  preferences.
\newblock In {\em Proceedings of the 2009 Conference on Empirical Methods in
  Natural Language Processing: Volume 2-Volume 2}, pages 638--647. Association
  for Computational Linguistics.

\bibitem[\protect\citename{Zhang and Nivre}2011]{zhang2011transition}
Yue Zhang and Joakim Nivre.
\newblock 2011.
\newblock Transition-based dependency parsing with rich non-local features.
\newblock In {\em Proceedings of the 49th Annual Meeting of the Association for
  Computational Linguistics: Human Language Technologies: short papers-Volume
  2}, pages 188--193. Association for Computational Linguistics.

\bibitem[\protect\citename{Zhou and Xu}2015]{zhou2015end}
Jie Zhou and Wei Xu.
\newblock 2015.
\newblock End-to-end learning of semantic role labeling using recurrent neural
  networks.
\newblock In {\em Proceedings of the Annual Meeting of the Association for
  Computational Linguistics}.

\end{thebibliography}

\end{document}